\documentclass[conference]{IEEEtran} \IEEEoverridecommandlockouts
\usepackage{cite} \usepackage{amsmath,amssymb,amsfonts}
\usepackage{algorithmic} \usepackage{graphicx}
\usepackage{textcomp} \usepackage{xcolor}
\usepackage{booktabs} \usepackage{multirow}
\usepackage{subcaption}  
\usepackage[
    colorlinks=false,      
    linkbordercolor=green, 
    citebordercolor=green, 
    pdfborder={0 0 1}      
]{hyperref}

\def\BibTeX{{\rm B\kern-.05em{\sc i\kern-.025em b}\kern-.08em
    T\kern-.1667em\lower.7ex\hbox{E}\kern-.125emX}}

\IEEEoverridecommandlockouts
\begin{document}

\title{Noise Aggregation Analysis Driven by Small-Noise Injection: Efficient Membership Inference for Diffusion Models}

\author{
\IEEEauthorblockN{Guo Li}
\IEEEauthorblockA{
\textit{South China University of Technology}\\
Guangzhou, China \\
csliguo@mail.scut.edu.cn
}
\and
\IEEEauthorblockN{Weihong Chen}
\IEEEauthorblockA{
\textit{South China University of Technology}\\
Guangzhou, China \\
cswilliam@mail.scut.edu.cn
}
\and
\IEEEauthorblockN{Yongfu Fan}
\IEEEauthorblockA{
\textit{University of Electronic Science and }\\
\textit{Technology of China}\\
yf.fan@std.uestc.edu.cn
}
}

\maketitle

\begin{abstract}
Diffusion models have demonstrated powerful performance in generating high-quality images. 
A typical example is text-to-image generator like Stable Diffusion. However, 
their widespread use also poses potential privacy risks. 
A key concern is membership inference attacks, 
which attempt to determine whether a particular data sample was used in the model training process.
Existing membership inference attacks against diffusion models either directly exploit sample loss differences 
or rely on image-level reconstruction differences.
Both approaches commonly ignore the consistency characteristics of noise prediction during the diffusion process, 
resulting in either low inference accuracy or high computational costs. 
To address these shortcomings, we propose a membership inference method based on noise aggregation analysis, 
and introduce a single-step, low-intensity noise injection diffusion strategy 
to amplify differences between member and non-member samples. 
Our proposed approach substantially reduces model query requirements while delivering more efficient and accurate membership inference.

\end{abstract}


\section{Introduction}
With the rapid development of generative models, high-quality image synthesis based on diffusion models has achieved significant 
success\cite{dhariwal2021diffusion,liu2023more,ramesh2022hierarchical,ruiz2023dreambooth,saharia2022photorealistic,rombach2022high}. 
Compared to GANs and VAEs, diffusion models demonstrate superior generalization and semantic consistency due to their unique training schemes~\cite{ho2022classifierfree,ma2023classifierguided}. However, this has also raised significant privacy concerns, specifically regarding membership inference attacks (MIA), which aim to determine whether a specific data sample was used during training~\cite{shokri2017membership,yeom2018privacy,salem2018ml,long2018understanding,long2020pragmatic}. For example, in medical scenarios, MIAs may result in the severe leakage of sensitive patient data; in commercial scenarios, competitors can exploit MIAs to identify proprietary training data. Consequently, investigating MIAs against diffusion models is particularly important for revealing internal mechanisms and facilitating the adoption of stronger privacy-preserving strategies.
\begin{figure}[t]
\centering
\includegraphics[width=0.49\textwidth]{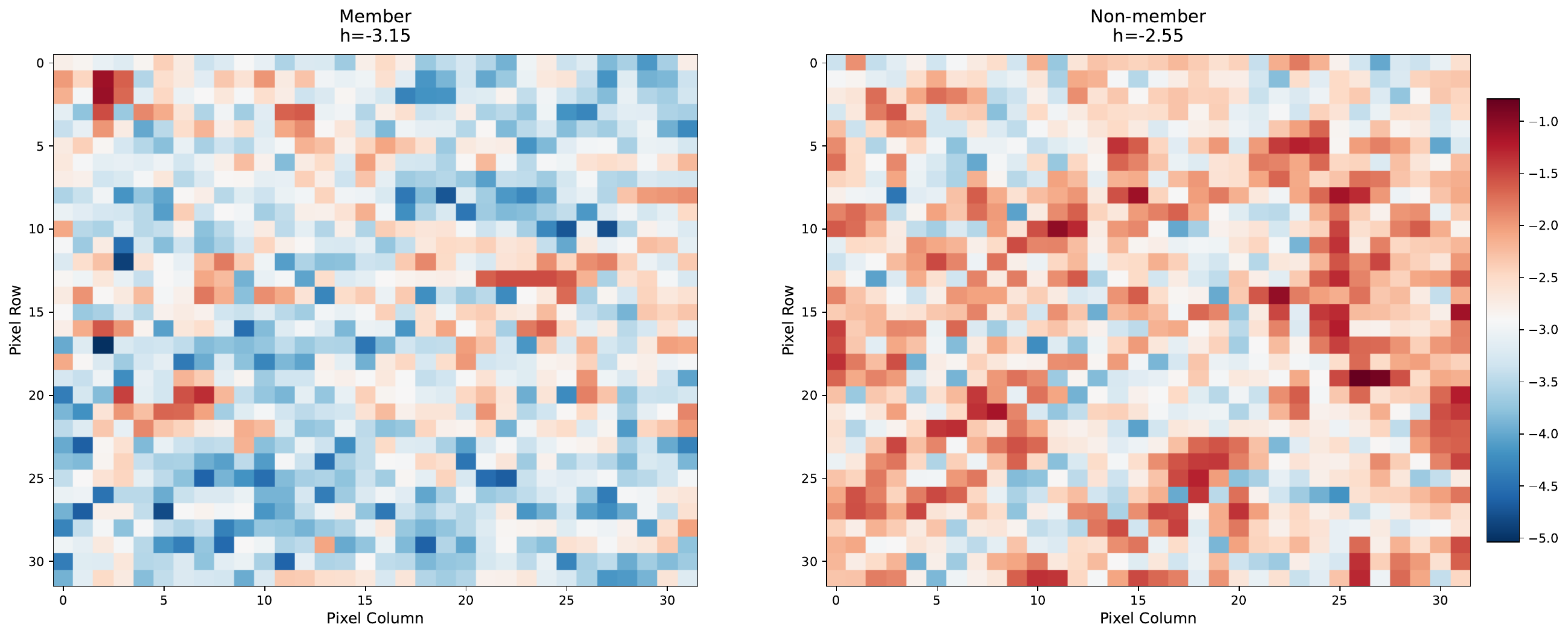}
\caption{
Conditional differential entropy maps of sample pairs. The left shows a member sample, and the right shows a non-member sample. 
The non-member sample exhibits noticeably redder regions, indicating substantially higher pixel-level uncertainty.}
\label{fig:2d_entropy}
\end{figure}

Existing MIA methods for diffusion models can be broadly categorized into three types. 
The first type leverages sample loss differences~\cite{matsumoto2023membership,hu2023membership}, 
assuming that member samples yield lower losses. 
But such methods introduce randomly sampled noise during loss computation. 
The randomness of the noise leads to instability in the attack results, which ultimately affects the overall attack performance.
The second type relies on image-level reconstruction and comparison to determine membership~\cite{duan2023diffusion,fu2025unlocking}, 
typically employing multi-step denoising strategies of DDIM deterministic sampling to mitigate the influence of random noise. 
However, this approach significantly increases the number of model queries, resulting in high computational cost.
The third category focuses on text-to-image diffusion models~\cite{zhai2024membership, li2024unveiling}, 
utilizing text features that limit their applicability to other types of diffusion architectures.

To overcome these limitations, we revisit diffusion models from a more fundamental perspective. Diffusion models can be naturally viewed as noise-prediction networks trained to estimate the injected noise at each diffusion timestep. From this perspective, overfitting in diffusion models is closely tied to overfitting in the underlying noise predictor, which may lead to memorization of training data. This observation motivates us to analyze the outputs of the noise-prediction network, providing a principled basis for designing effective membership inference attacks against diffusion models.
By comparing the pixel-level conditional differential entropy heatmaps of member and non-member samples within a local neighborhood of specific timesteps 
(as shown in Figure~\ref{fig:2d_entropy}), we observe that non-member samples exhibit substantially higher uncertainty.
To further quantify this uncertainty and achieve more accurate membership inference, 
we introduce a metric method based on noise aggregation that measures the consistency of predicted noise vectors 
across a sequence of denoising steps around a specific timestep. 
Our intuition is that, since member samples participate in the training process, 
The noise-prediction network generates more accurate noise estimates for member samples, 
and consequently shows stronger consistency in its predictions across the neighborhood of specific timesteps.
In contrast, predictions for non-member samples exhibit stronger randomness due to their absence from the training process.

\begin{figure*}[t] \centering
\includegraphics[width=1\textwidth]{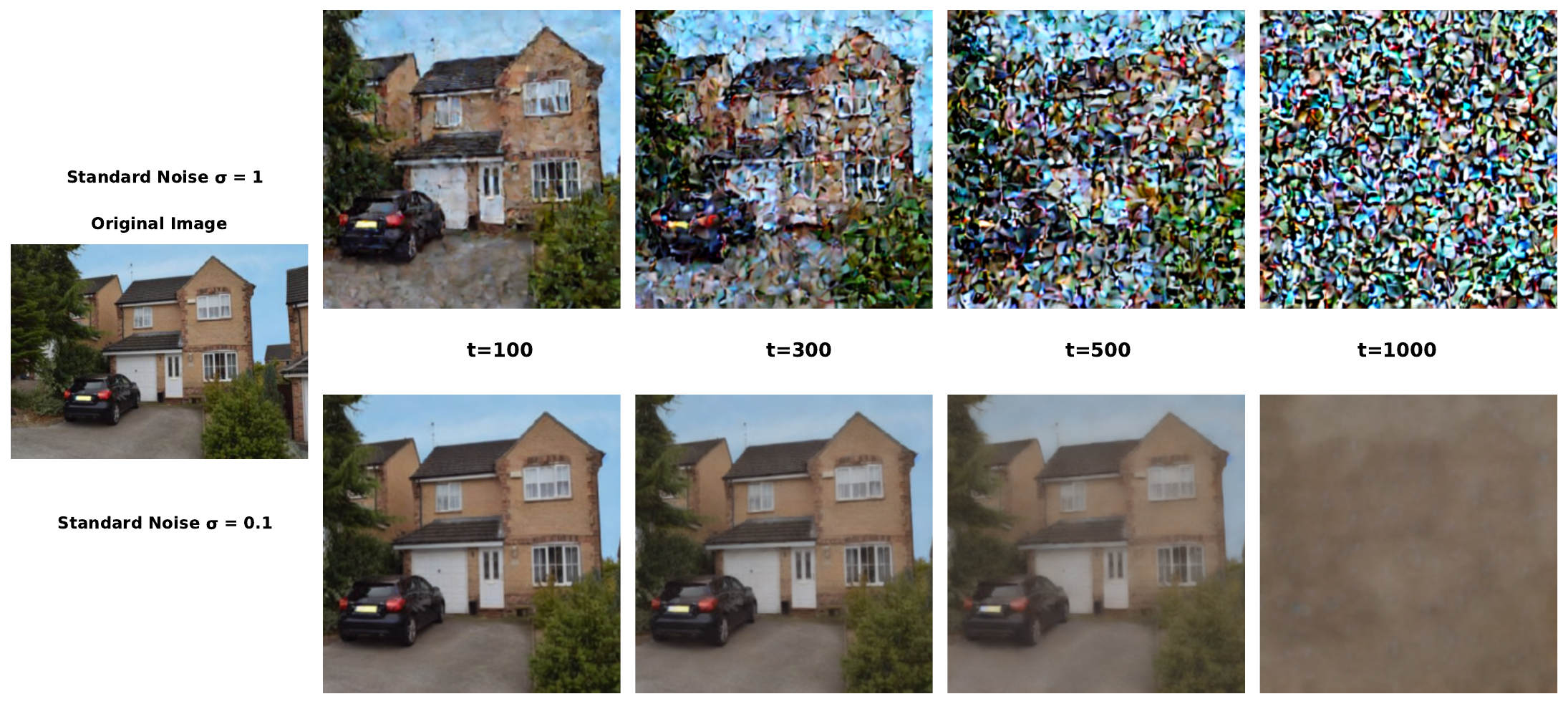}
\caption{Comparison of diffusion effects under different noise intensities. 
The first row shows the diffusion process after injecting noise with standard deviation $\sigma = 1$, 
while the second row shows the process with standard deviation $\sigma = 0.1$. 
The results indicate that low-intensity noise better preserves the overall image structure and 
is more favorable for subsequent membership inference.
} \label{fig:noise_comparison}
\end{figure*}

We additionally observe that high-intensity noise injection disrupts semantic information in images (as shown in Figure~\ref{fig:noise_comparison}), 
significantly degrading inference performance. 
Conversely, low-intensity noise better preserves structural content and amplifies inter-class discrepancies 
between member and non-member samples. 
Building on this insight, we adopt a single-step small-noise injection strategy to approximate one diffusion step toward the target timestep, 
which not only drastically reduces the number of model queries but also improves membership inference accuracy.

In this way, we propose a membership inference method based on noise aggregation analysis and small-noise injection strategy.
To validate the effectiveness of our method, 
we conducted extensive experiments. 
The results demonstrate that our approach achieves comprehensive improvements on DDPM benchmarks and also delivers strong performance on text-to-image models. 

\begin{figure*}[t]
\centering
\includegraphics[width=1\textwidth]{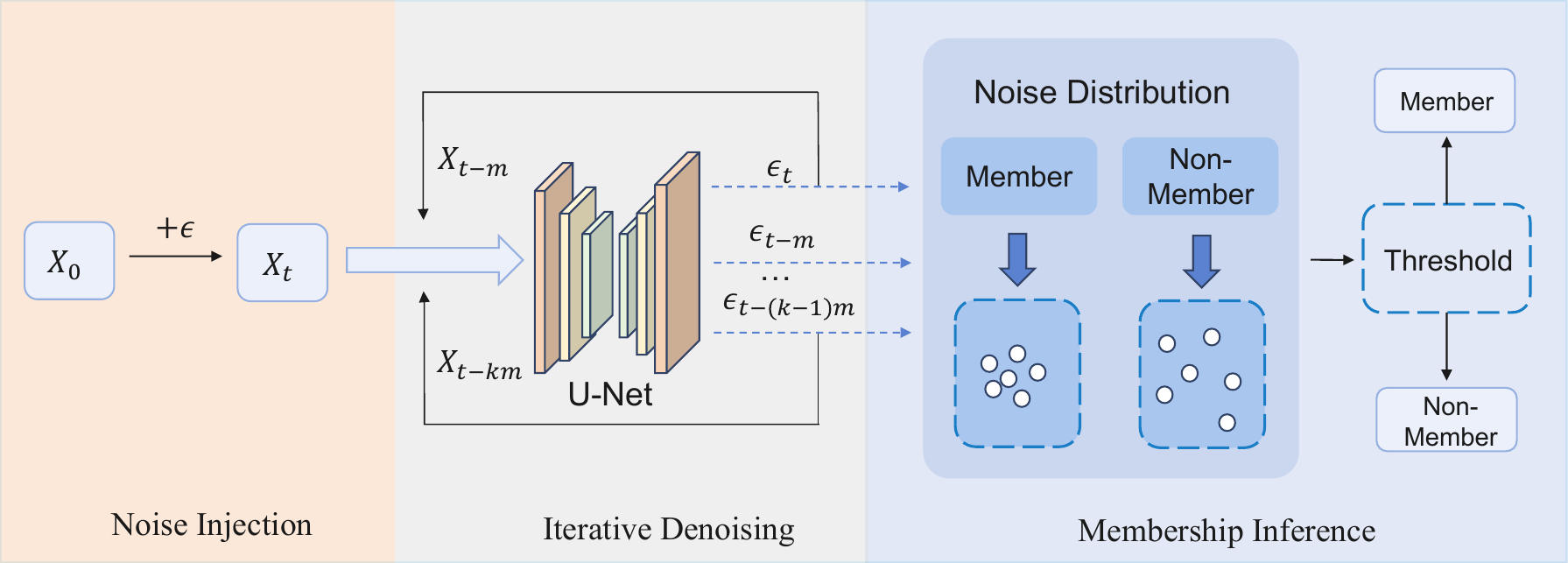}
\caption{Overview of our proposed membership inference attack pipeline. 
The approach injects small noise $\epsilon$ into test images $X_0$, 
predicts noise at selected timesteps $t$, 
and evaluates the aggregation degree of predictions to determine the membership of the sample.}
\label{fig:pipeline}
\end{figure*}

The main contributions of this paper are summarized as follows:
\begin{itemize}
  \item  To the best of our knowledge, we are the first to explicitly leverage noise aggregation analysis for membership inference attack against diffusion models. 
  We argue that noise sampling in diffusion models is inherently stochastic, 
  and analyzing noise aggregation can effectively mitigate this randomness, enabling more accurate membership inference.
  \item We introduce an efficient diffusion-based sampling strategy for membership inference attacks based on one-time small-noise injection. 
This strategy substantially reduces sampling cost while amplifying the inter-class differences between members and non-members, leading to improved attack efficiency and effectiveness.
  \item   We conduct extensive experiments on multiple datasets to validate the effectiveness of the proposed method.
Through ablation studies and sample distribution analysis, we identify the key factors that are critical to the attack performance and provide insights into the memorization behavior of diffusion models.
\end{itemize}

\section{Related Work}
\subsection{Diffusion Models}
Diffusion models~\cite{ho2020denoising, sohl2015deep, nichol2021improved} have emerged as the leading approach for high-quality image generation, offering superior stability and quality compared to GANs~\cite{goodfellow2014generative} and VAEs~\cite{kingma2013auto} through their iterative denoising process. Beyond image synthesis, their flexible conditional control has revolutionized diverse fields, including video~\cite{bar2024lumiere,ma2024latte}, speech~\cite{popov2021grad,jeong2021diff,long2025mixdiff}, and molecular design~\cite{huang2024dual,oestreich2025drugdiff}.
At the same time, researchers have proposed many improvements to increase efficiency and quality of the diffusion models, such as accelerated sampling (DDIM~\cite{song2020denoising}), latent space diffusion (LDM~\cite{rombach2022high}), and architectural optimizations that combine attention mechanisms with large-scale pre-training\cite{rombach2022high,ramesh2022dall,saharia2022photorealistic}. 
These advances have greatly expanded the application scope of diffusion models and raised broad discussions on privacy, security, and copyright. 

\subsection{Membership Inference Attacks}
Membership Inference Attacks (MIA) aim to determine whether a specific sample was used to train a target model. Initially focused on classification tasks, early works established foundational strategies using shadow models and loss-based metrics~\cite{shokri2017membership,salem2018ml,yeom2018privacy,long2018understanding,long2020pragmatic}. As generative models gained popularity, research expanded to GANs and VAEs, where scholars developed white-box and black-box attacks based on generator outputs, Monte Carlo integration, and reconstruction errors~\cite{hayes2017logan,hilprecht2019monte,liu2019performing,chen2020gan}. These studies revealed fundamental privacy vulnerabilities in generative tasks, laying the groundwork for investigating more complex architectures.

With the widespread adoption of diffusion models, researchers have adapted MIA strategies to this domain, employing techniques such as posterior estimation, pixel error analysis, and likelihood comparison~\cite{duan2023diffusion,matsumoto2023membership,hu2023membership,zhai2024membership,li2024unveiling}. However, existing methods exhibit notable limitations: loss-based approaches often fail to account for sampling stochasticity, limiting accuracy~\cite{matsumoto2023membership,hu2023membership}; reconstruction-based methods incur high computational costs due to extensive model queries~\cite{duan2023diffusion,fu2025unlocking}; and text-conditioned strategies often struggle to generalize across architectures~\cite{zhai2024membership,li2024unveiling}. 
To address these challenges, we propose a method 
that leverages the consistency of noise predictions across adjacent timesteps 
to analyze the aggregation of predicted noise, 
effectively mitigating the instability caused by random noise sampling. 
Meanwhile, we introduce a single-step small-noise injection strategy 
that significantly reduces the number of model queries, thereby improving the overall efficiency of the attack.

\section{Preliminaries}
Denoising Diffusion Probabilistic Models (DDPM~\cite{ho2020denoising}) define a stochastic Markov 
chain that gradually transforms an image into noise. The forward process, 
denoted as $q$, iteratively adds Gaussian noise to the data at each timestep. 
This process can be written as:
\begin{equation}
q(x_{1:T} | x_0) = \prod_{t=1}^T q(x_t | x_{t-1})
\end{equation}
\begin{equation}
q(x_t|x_{t-1}) = \mathcal{N}(x_t; \sqrt{1-\beta_t}x_{t-1}, \beta_t I)
\label{eq:q_forward}
\end{equation}
where $\beta_t$ is a variance schedule that controls the noise level at step $t$.
As $t$ increases, $\bar{\alpha}_t$ approaches zero, making $x_t$ isotropic
Gaussian noise. The reverse process aims to reconstruct the data distribution
from Gaussian noise. This process can be expressed as:
\begin{equation}
p(x_{0:T}) = p(x_T) \prod_{t=1}^T p_\theta(x_{t-1}|x_t)
\end{equation}
\begin{equation}
p_\theta(x_{t-1}|x_t) = \mathcal{N}(x_{t-1}; \mu_\theta(x_t, t), \Sigma_\theta(x_t, t))
\label{eq:p_reverse}
\end{equation}
where $\Sigma_\theta(x_t, t)$ is a constant depending on the variance schedule $\beta_t$, and
$\mu_\theta(x_t, t)$ is determined by a neural network. Through recursive
application of reverse steps, Gaussian noise is converted back to the original
image.

Training DDPM requires sampling image $x_0$, timestep $t$, and random noise
$\epsilon \sim \mathcal{N}(0, I)$. The forward process generates noisy image
$x_t$. A U-Net $\epsilon_\theta$ predicts the noise in $x_t$. The loss function
for the denoising U-Net is:
\begin{equation}
L = \mathbb{E}_{x_0, t, \epsilon}[\|\epsilon - \epsilon_\theta(x_t, t)\|^2]
\label{eq:loss}
\end{equation}

To control the randomness of the reverse process, DDIM~\cite{song2020denoising} modifies the noise added
at each step:
\begin{equation} x_{t-1} = \sqrt{\bar{\alpha}_{t-1}}\hat{x}_0 +
\sqrt{1-\bar{\alpha}_{t-1}-\sigma_t^2}\epsilon_\theta(x_t, t) + \sigma_t\epsilon_t
\label{eq:ddim}
\end{equation}
where $\hat{x}_0$ is the estimated initial data, with expressions:
\begin{equation}
\hat{x}_0 = \frac{x_t - \sqrt{1-\bar{\alpha}_t}\epsilon_\theta(x_t,
t)}{\sqrt{\bar{\alpha}_t}} \end{equation}

\section{Methodology}
This section provides a detailed exposition of our proposed membership inference
attack. As illustrated in
Fig.~\ref{fig:pipeline}, our approach consists of the following three stages: small-scale
noise injection, iterative denoising prediction, and noise aggregation degree quantification analysis. 
Overall, we infer the membership by injecting small-noise into the image and 
then quantifying the spatial aggregation degree of the predicted noise.

\subsection{Small-Scale Noise Injection Strategy}

The forward process of diffusion models is inherently a stochastic Markov chain that gradually adds Gaussian noise to the data. Based on the reparameterization trick, the cumulative noise injection from the initial state $x_0$ to any timestep $t$ can be mathematically equivalent to a single-step transformation:
\begin{equation}
x_t = \sqrt{\bar{\alpha}_t} x_0 + \sqrt{1 - \bar{\alpha}_t} \epsilon
\label{eq:xt_direct}
\end{equation}
where $\epsilon \sim \mathcal{N}(0, I)$ is standard Gaussian noise. This property allows us to directly obtain noisy images at target timesteps without iterative computation.

However, directly sampling with standard large variance poses limitations for membership inference. As shown in the first row of Fig.~\ref{fig:noise_comparison}, standard noise ($\sigma=1$) introduces severe high-frequency corruption. Even at early timesteps, the semantic structure of the original image is rapidly obliterated and transformed into unrecognizable random noise. This destruction pushes the sample entirely out of the model's local memorization scope, compressing the statistical distance between members and non-members.

To address this, we propose a strategy based on small-scale noise injection. Our motivation draws on the geometry of the loss landscape, particularly the distinction between sharp and flat minima~\cite{keskar2016large}. We hypothesize that member samples, having been overfitted, typically reside in sharp minima (steep basins of attraction), whereas non-members are located in flatter regions. 
As demonstrated in the second row of Fig.~\ref{fig:noise_comparison}, applying a limited noise variance (e.g., $\sigma=0.1$) acts as a structure-preserving perturbation. Unlike the standard process, the image content remains visually recognizable, indicating the sample is still retained within its potential attraction domain. For member samples, the steep gradients characteristic of sharp minima exert a consistent "restoration force" on these preserved structures, pulling the prediction back to the origin. Conversely, the recovery of non-members exhibits higher uncertainty. This approach amplifies the distinguishability of members while maintaining high computational efficiency.
\subsection{Iterative Denoising Prediction Stage}
After obtaining the noisy image $x_t$, we acquire noise predictions at different
timesteps through an iterative denoising process to construct feature vectors
for membership analysis. This process is based on the intuition that
training set members, having been repeatedly optimized during model training,
exhibit higher consistency and concentration in the model's noise predictions.

First, we input the noisy image $x_t$ into the target U-Net denoising network to
obtain the predicted noise at timestep $t$:
\begin{equation} \hat{\epsilon}_t = \epsilon_\theta(x_t, t)
\end{equation}
Based on the predicted noise, we can estimate the original image:
\begin{equation} \hat{x}_0 = \frac{x_t -
\sqrt{1-\bar{\alpha}_t}\hat{\epsilon}_t}{\sqrt{\bar{\alpha}_t}}
\end{equation}
Based on Eq.~\eqref{eq:ddim}, we set $\sigma_t = 0$, the process of denoising becomes deterministic.
Then we can estimate the original image at timestep $t-m$:
\begin{equation} x_{t-m} = \sqrt{\bar{\alpha}_{t-m}}\hat{x}_0 +
\sqrt{1-\bar{\alpha}_{t-m}}\hat{\epsilon}_t
\end{equation}
where $m$ is the stride between sampled timesteps. By repeating the above denoising
process, we sequentially obtain predicted noise sequences for $k$ consecutive
timesteps
$\{\hat{\epsilon}_{t}, \hat{\epsilon}_{t-m}, \hat{\epsilon}_{t-2m}, \ldots,
\hat{\epsilon}_{t-(k-1)m}\}$.
This noise sequence constitutes the fundamental feature representation for
subsequent membership analysis.

\begin{table*}[t]
\centering
\caption{Comparison of ASR and AUC across different datasets. 
\textbf{Bold} values denote the best results, and \textcolor{blue}{blue} values denote the second-best results.
The $\uparrow$ symbol means higher is better.}
\begin{tabular}{@{}
l@{\hspace{1.2em}}c@{\hspace{1.2em}}
c@{\hspace{1.5em}}c@{\hspace{1.2em}}
c@{\hspace{1.5em}}c@{\hspace{1.2em}}
c@{\hspace{1.5em}}c@{\hspace{1.2em}}
c@{\hspace{1.5em}}c@{}}
\toprule
 &  &
\multicolumn{2}{c}{CIFAR-10} &
\multicolumn{2}{c}{CIFAR-100} &
\multicolumn{2}{c}{Tiny-IN} &
\multicolumn{2}{c}{Average} \\
\cmidrule(lr){3-4} 
\cmidrule(lr){5-6} 
\cmidrule(lr){7-8} 
\cmidrule(lr){9-10}
Method & Query & 
\multicolumn{1}{c@{\hspace{2.2em}}}{ASR$\uparrow$} & AUC$\uparrow$ &
\multicolumn{1}{c@{\hspace{2.2em}}}{ASR$\uparrow$} & AUC$\uparrow$ &
\multicolumn{1}{c@{\hspace{2.2em}}}{ASR$\uparrow$} & AUC$\uparrow$ &
\multicolumn{1}{c@{\hspace{2.2em}}}{ASR$\uparrow$} & AUC$\uparrow$ \\
\midrule
GAN-Leaks & 1000 & 0.615 & 0.646 & 0.513 & 0.459 & 0.545 & 0.457 & 0.558 & 0.521 \\
NaiveLoss & 1 & 0.663 & 0.718 & 0.654 & 0.709 & 0.646 & 0.700 & 0.654 & 0.709 \\
SecMI & 12 & \textcolor{blue}{0.811} & \textcolor{blue}{0.881} & \textcolor{blue}{0.798} & \textcolor{blue}{0.868} & \textcolor{blue}{0.821} & \textcolor{blue}{0.894} & \textcolor{blue}{0.810} & \textcolor{blue}{0.881} \\
\midrule
Ours & 5 &
\textbf{0.901} & \textbf{0.957} &
\textbf{0.839} & \textbf{0.903} &
\textbf{0.842} & \textbf{0.912} &
\textbf{0.861} & \textbf{0.924} \\
\bottomrule
\end{tabular}
\label{tab:main_results}
\end{table*}

\begin{table*}[t]
\centering
\caption{Comparison of TPR @ 1\% FPR (\%) and TPR @ 0.1\% FPR (\%) across different datasets. 
\textbf{Bold black} values denote the best results, and \textcolor{blue}{blue} values denote the second-best results. 
The $\uparrow$ symbol means higher is better.}
\begin{tabular}{@{}l@{\hspace{1.5em}}
cc@{\hspace{1.5em}}
cc@{\hspace{1.5em}}
cc@{}}
\toprule
 &
\multicolumn{2}{c}{CIFAR-10} &
\multicolumn{2}{c}{CIFAR-100} &
\multicolumn{2}{c}{Tiny-IN} \\
\cmidrule(lr){2-3}
\cmidrule(lr){4-5}
\cmidrule(lr){6-7}
Method &
TPR@1\%$\uparrow$ & TPR@0.1\%$\uparrow$ &
TPR@1\%$\uparrow$ & TPR@0.1\%$\uparrow$ &
TPR@1\%$\uparrow$ & TPR@0.1\%$\uparrow$ \\
\midrule
GAN-Leaks  & 2.80 & 0.29 & 1.85 & 0.23 & 1.01 & 0.13 \\
NaiveLoss  & 3.35 & 0.40 & 4.79 & \textcolor{blue}{0.76} & 4.44 & 0.44 \\ 
SecMI      & \textcolor{blue}{9.11} & \textcolor{blue}{0.66} & \textcolor{blue}{9.26} & 0.46 & \textcolor{blue}{12.67} & \textcolor{blue}{0.96} \\
\midrule
Ours &
\textbf{28.7} & \textbf{1.22} &
\textbf{9.65} & \textbf{0.78} &
\textbf{14.58} & \textbf{1.03} \\
\bottomrule
\end{tabular}
\label{tab:tpr_results}
\end{table*}

\begin{figure*}[t]
  \centering
  \begin{minipage}[b]{0.49\textwidth}
      \centering
      \begin{subfigure}[b]{0.49\textwidth}
          \centering
          \includegraphics[width=\textwidth]{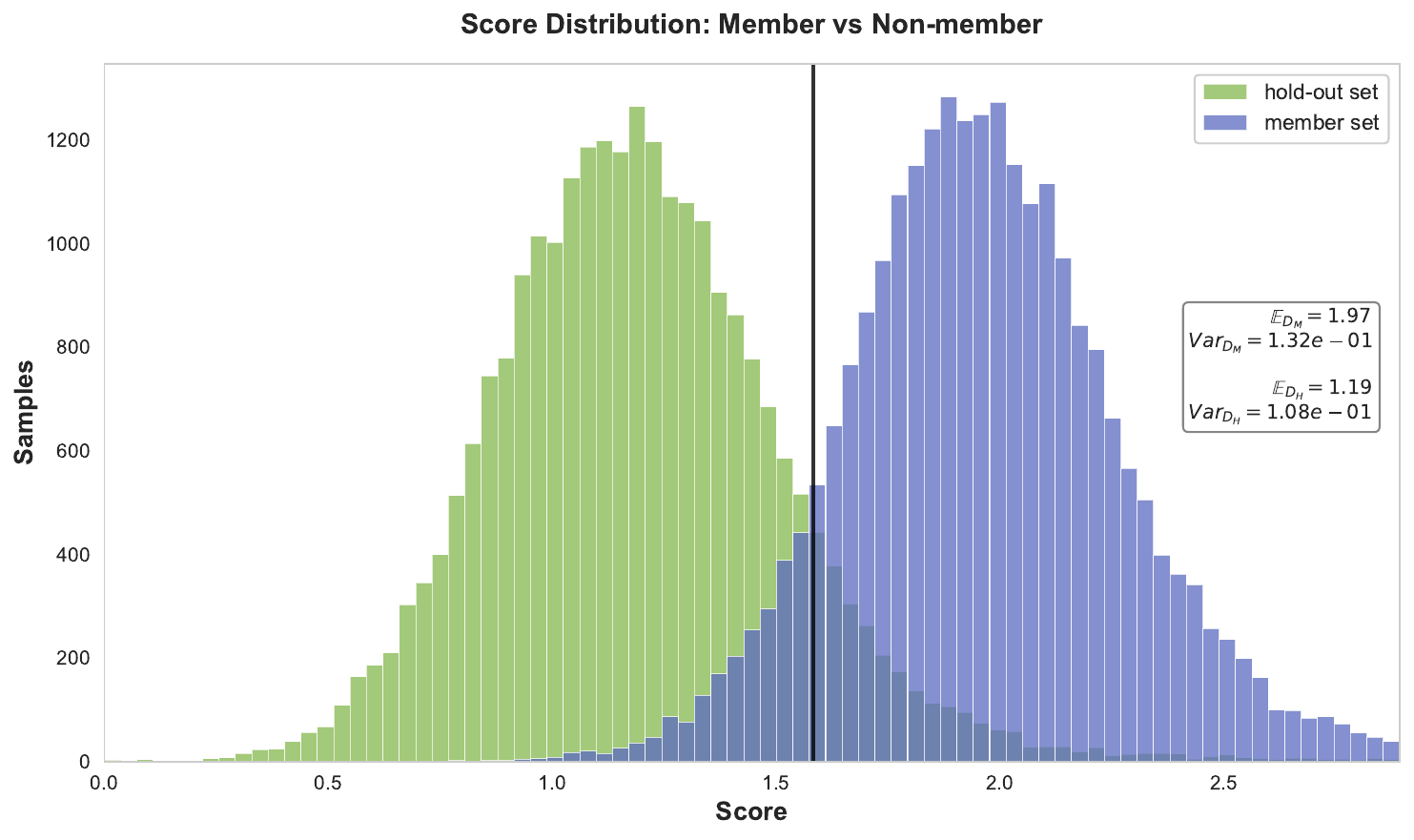}
          \caption{CIFAR10}
      \end{subfigure}
      \hfill
      \begin{subfigure}[b]{0.49\textwidth}
          \centering
          \includegraphics[width=\textwidth]{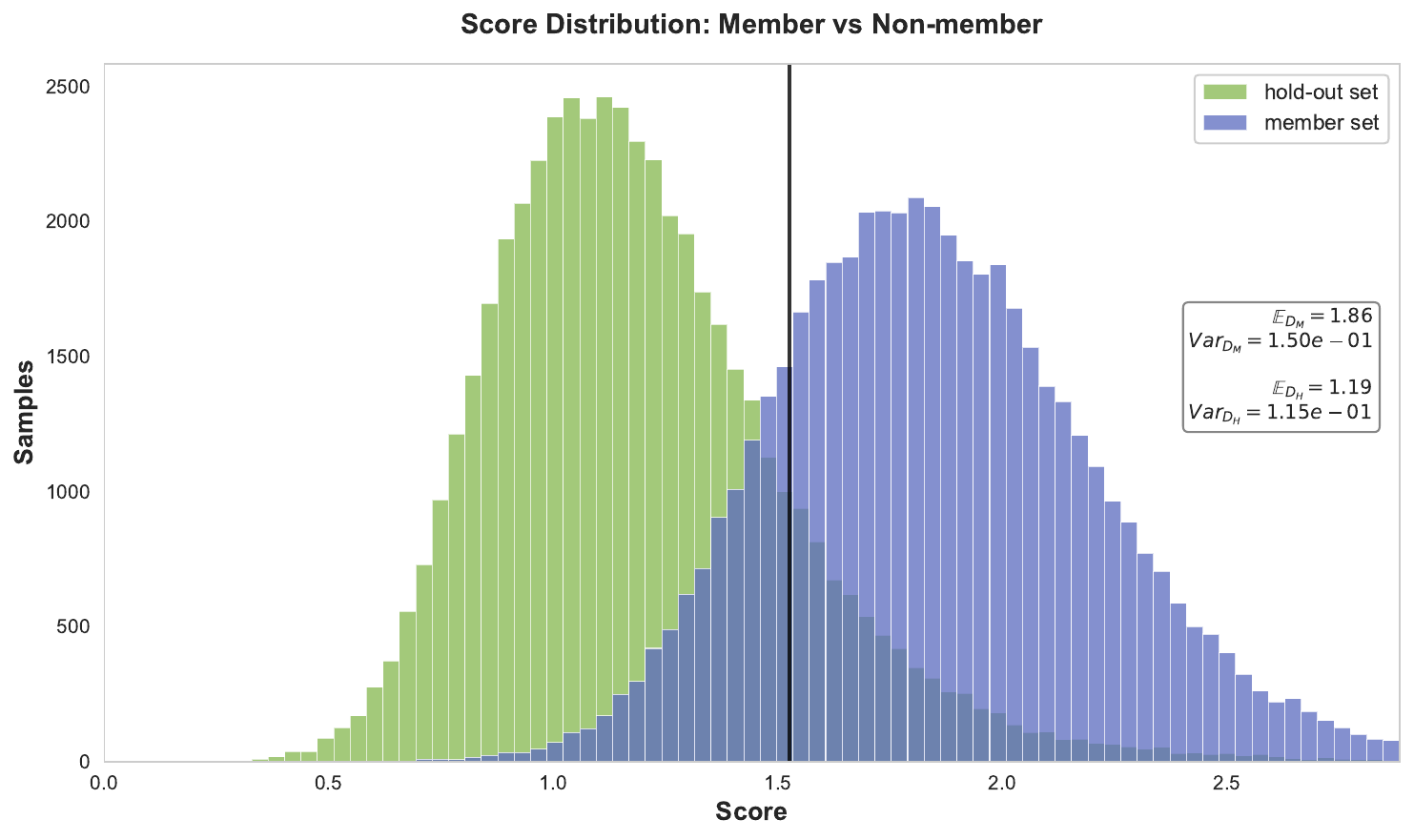}
          \caption{Tiny-IN}
      \end{subfigure}
      \caption{Distribution histogram of member samples and non-member samples}
      \label{fig:histogram_comparison}
  \end{minipage}
  \hfill
  \begin{minipage}[b]{0.49\textwidth}
      \centering
          \includegraphics[width=\textwidth]{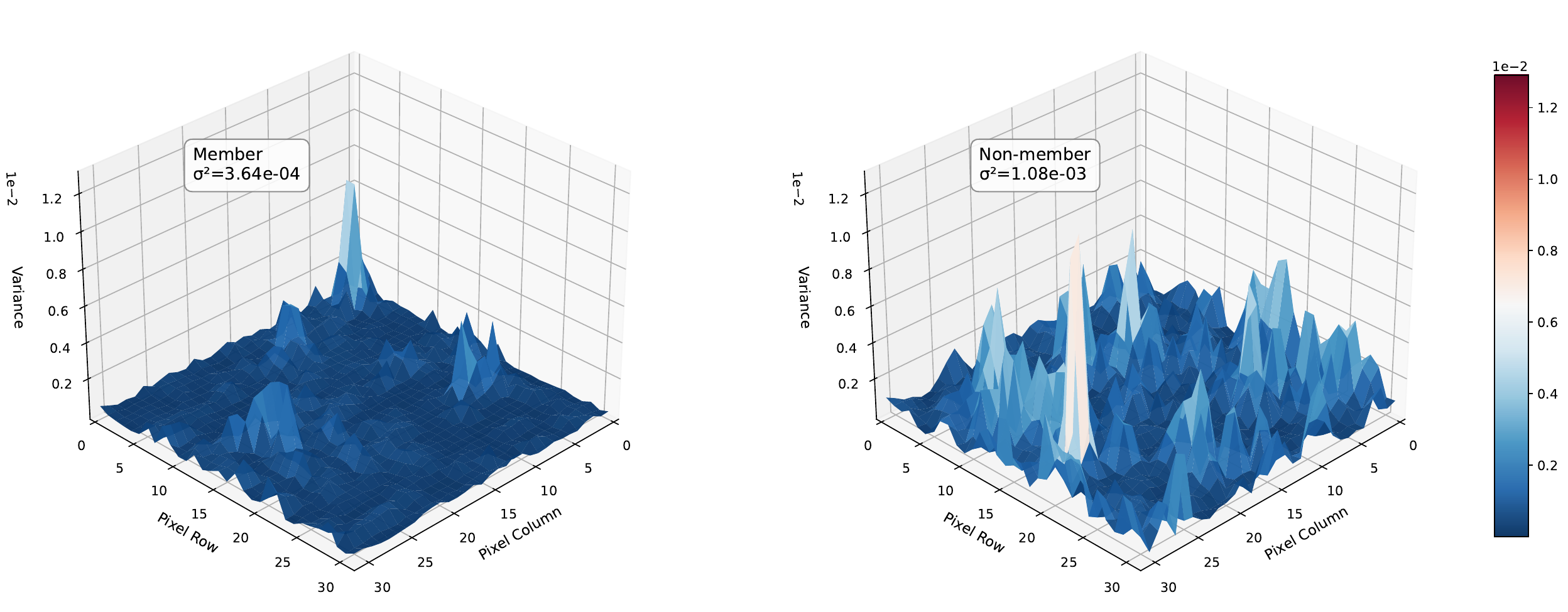}
      \caption{3D heat map of variance between member samples and non-member samples}
      \label{fig:3d_heatmap_comparison}
  \end{minipage}
\end{figure*}

\begin{table*}[htbp]
\centering
\begin{minipage}{0.48\textwidth}
  \centering
  \caption{Attack performance across denoising steps}
  \label{tab:k_performance}
  \begin{tabular}{c@{\hspace{1.2em}}c@{\hspace{1.2em}}ccc}
  \toprule
  K & ASR$\uparrow$ & AUC$\uparrow$ & TPR@1\%FPR$\uparrow$ & TPR@0.1\%FPR$\uparrow$ \\
  \midrule
  2 & 0.888 & 0.944 & 25.5 & 1.19 \\
  3 & 0.892 & 0.952 & 26.9 & 1.18 \\
  4 & 0.897 & 0.955 & 29.6 & 1.20 \\
  5 & 0.901 & 0.956 & 29.7 & 1.28 \\
  6 & 0.900 & 0.956 & 26.9 & 1.01 \\
  7 & 0.893 & 0.952 & 22.0 & 0.74 \\
  8 & 0.875 & 0.938 & 17.7 & 0.63 \\
  \bottomrule
  \end{tabular}
  \label{tab:denoising_steps}
\end{minipage}
\hfill
\begin{minipage}{0.48\textwidth}
  \centering
  \caption{Attack performance across initial noise intensity }
  \label{tab:latent_diffusion}
  \begin{tabular}{ccccc}
    \toprule
    std\_dev & ASR$\uparrow$ & AUC$\uparrow$ & TPR@1\%FPR$\uparrow$ & TPR@0.1\%FPR$\uparrow$ \\
    \midrule
    0.05 & 0.884 & 0.944 & 16.20 & 0.864 \\
    0.10 & 0.901 & 0.957 & 30.10 & 0.980 \\
    0.15 & 0.896 & 0.955 & 31.20 & 2.170 \\
    0.20 & 0.882 & 0.945 & 24.80 & 1.880 \\
    0.30 & 0.862 & 0.927 & 14.50 & 0.696 \\
    0.50 & 0.837 & 0.903 & 7.73 & 0.452 \\
    1.00 & 0.742 & 0.806 & 4.12 & 0.292 \\
  \bottomrule
  \end{tabular}
  \label{tab:noise_intensity_analysis}
\end{minipage}
\end{table*}

\subsection{Noise Aggregation Degree Quantification Analysis}
From an information-theoretic perspective, member samples are repeatedly optimized during training, 
resulting in more concentrated predicted noise and lower entropy. 
In contrast, non-member samples exhibit greater uncertainty in their predicted noise, 
leading to higher entropy and more dispersed predictions. 
Let $( H(\epsilon | x) )$ denote the conditional differential entropy of the predicted noise given a sample ( x ). 
The above hypothesis can therefore be expressed as
\begin{equation}
H(\epsilon|x_{member}) < H(\epsilon|x_{non-member})
\end{equation}
As shown in Fig.~\ref{fig:2d_entropy}, we visualize the differential entropy of member and non-member samples using 2D heatmaps. 
It can be observed that the heatmaps of non-member samples appear  redder, 
indicating higher entropy values and greater uncertainty compared to member samples.
To further quantify the uncertainty between member and non-member samples, 
we introduce the noise aggregation degree to characterize the consistency of the predicted noise.

Let the noise sequence be denoted as $\mathcal{E} = \{\hat{\epsilon}_{t-im}\}_{i=0}^{k-1}$,
and define the noise aggregation degree as $C(\mathcal{E})$.
We measure $C(\mathcal{E})$ by computing the average pairwise Euclidean distance among all predicted noises, formulated as:
\begin{equation}
C(\mathcal{E}) = \frac{1}{|\mathcal{E}|^2} \sum_{i,j} \|\hat{\epsilon}_i - \hat{\epsilon}_j\|_2
\end{equation}
To enhance numerical stability, the final membership score is defined as:
\begin{equation} S_m = -\log(C(\mathcal{E}) + \delta)
\end{equation}
where $\delta$ is a numerical stability constant. 
Then the membership inference function is defined as:
\begin{equation}
\mathcal{A}(x, g_\phi) = \mathbb{1}[S_m \geq \tau]
\end{equation}

where $\mathbb{1}$ denotes the indicator function, which outputs 1 if the membership score exceeds the threshold $\tau$,
indicating that the sample is inferred as a member;

\section{Experiments}
\subsection{Experimental Setup}

\textbf{Dataset Configuration:} We conducted comprehensive evaluations on three
widely used visual datasets: CIFAR-10, CIFAR-100, and Tiny-ImageNet (Tiny-IN). We
employed a random partitioning strategy, allocating 50\% of samples from each
dataset as the training set and the remaining 50\% for testing. Specifically,
the training and test sets for CIFAR-10 and CIFAR-100 each contain 25,000
samples, while Tiny-IN contains 50,000 samples each.

For text-to-image model evaluation, as they trained on LAION
datasets, we randomly sampled 1,000 images from it as the member set.
For the non-member set, we randomly sampled 1,000 images from the COCO2017-Val dataset, 
which is commonly used for generative model evaluation.

\textbf{Implementation Details:} The training procedures and hyperparameter
settings for diffusion models remain consistent with~\cite{duan2023diffusion} to
ensure fair comparison. Key parameter settings include: attack diffusion steps $t=80$,
noise prediction sampling count $k=5$, and DDIM accelerated sampling denoising
steps $m=10$. For text-to-image models, we directly employ pre-trained Stable
Diffusion v1.4 and v1.5 models from HuggingFace as attack targets.

\textbf{Evaluation Metric:} 
We adopt standard evaluation metrics widely used in previous membership inference research, 
including Attack Success Rate (ASR), Area Under the ROC Curve (AUC), 
and True Positive Rate at fixed false positive rates.

\subsection{Baseline Method Comparison}
We compare against a suite of representative baselines, 
including GAN-Leaks~\cite{chen2020gan}, NaiveLoss~\cite{matsumoto2023membership}, and SecMI~\cite{duan2023diffusion}. 
To ensure a fair evaluation, we adopt a unified protocol based on a threshold decision rule without training on the test set.
And we follow the implementation details reported in each method’s original paper.
We trained DDPM models on CIFAR-10, CIFAR-100, and Tiny-IN 
for comprehensive performance comparison.

As shown in Table~\ref{tab:main_results} and Table~\ref{tab:tpr_results}, 
our method outperforms all existing approaches on all metrics.
Moreover, compared with SecMI, which achieves the best overall performance among existing approaches, 
our method significantly reduces the number of model queries, 
thereby greatly improving the efficiency of membership inference attacks.

\subsection{Comparative Analysis of Member and Non-member Samples}
As illustrated in Figure~\ref{fig:histogram_comparison}, 
we plot the histogram of membership scores obtained using our proposed method. 
It can be observed that the score distributions of member and non-member samples 
differ significantly. Member samples generally exhibit higher scores, 
which is the key factor enabling the effectiveness of our approach.

To further provide an intuitive visualization of the differences between member and non-member samples, 
we also plot their corresponding 3D heatmaps, 
which display the variance of each pixel’s predicted noise within the neighborhood of a specific timestep. 
As shown in Figure~\ref{fig:3d_heatmap_comparison}, 
the member samples appear darker in color, 
indicating smaller pixel-wise variations and a higher degree of noise aggregation. 
This demonstrates that member samples exhibit stronger consistency in their predicted noise during the generative process.

\begin{table}[t] \centering
\caption{Impact of different aggregation degree metrics}
\begin{tabular}{@{}lcccc@{}}
\toprule
Aggregation Metrics & ASR$\uparrow$ & AUC$\uparrow$ & TPR@1\%FPR$\uparrow$
(\%)\\ \midrule
Distance to centroid & 0.899 & 0.956 & 28.5 \\ 
Convex hull volume & 0.771 & 0.839 & 5.9 \\
Average density & 0.901 & 0.957 & 28.2\\ 
L1 average distance & 0.885 & 0.942 & 12.9 \\ 
L2 average distance & 0.901 & 0.957 & 29.7 \\ \bottomrule
\end{tabular} 
\label{tab:aggregation_metrics}
\end{table}

\subsection{Ablation Studies}

\subsubsection{Denoising Steps Impact}
Our method relies on analyzing the aggregation of predicted noises across multiple consecutive timesteps, 
making the choice of the number of noise samples crucial. 
To investigate this, we conducted experiments on the CIFAR-10 dataset, 
with results shown in Table~\ref{tab:denoising_steps}. 
As the number of denoising steps K increases, 
the performance of our method exhibits a “rise-then-fall” trend. 
We think this is because when the number of denoising steps is small, 
the inherent randomness of the sampling noise dominates, 
preventing the aggregation analysis from fully demonstrating its advantage.
And when the number of denoising steps is large, 
the denoising process introduces noticeable differences between images, 
which amplifies the differences in predicted noise 
and leads to a performance drop. 
Notably, even with $K=2$, our method outperforms the current state-of-the-art method~\cite{duan2023diffusion}, 
further demonstrating that our method achieves efficient inference while significantly reducing the cost of the attack.

\subsubsection{Initial Noise Intensity Impact Analysis}
Our method enhances attack efficiency by injecting low-intensity noise and performing a single-step diffusion, 
which greatly reduces the number of model queries.
As shown in Table~\ref{tab:noise_intensity_analysis}, 
we further investigate the impact of different noise standard deviations on attack performance. 
The attack achieves optimal performance when the standard deviation of the injected noise is around 0.1.
When the noise variance is too small or too large, 
the attack effectiveness decreases. 
We think this is because when the noise level is too low, 
the image maintains a high signal-to-noise ratio, 
resulting in a lack of clear distinction between member and non-member samples.
When the noise level is too high, it destroys the semantic information of the image, 
making it difficult even for member samples to accurately predict the noise, 
which weakens the distinction between member and non-member samples.

\begin{figure*}[t]
    \begin{minipage}[b]{0.49\textwidth}
        \centering
        \begin{subfigure}[b]{0.49\textwidth}
            \centering
            \includegraphics[width=\textwidth]{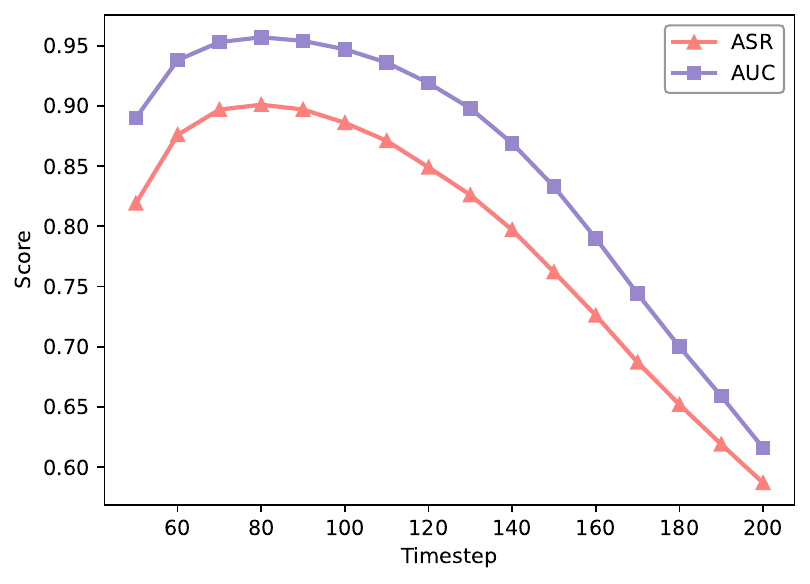}
            \caption{ASR and AUC}
            \label{fig:timestep_asr_auc}
        \end{subfigure}
        \hfill
        \begin{subfigure}[b]{0.49\textwidth}
            \centering
            \includegraphics[width=\textwidth]{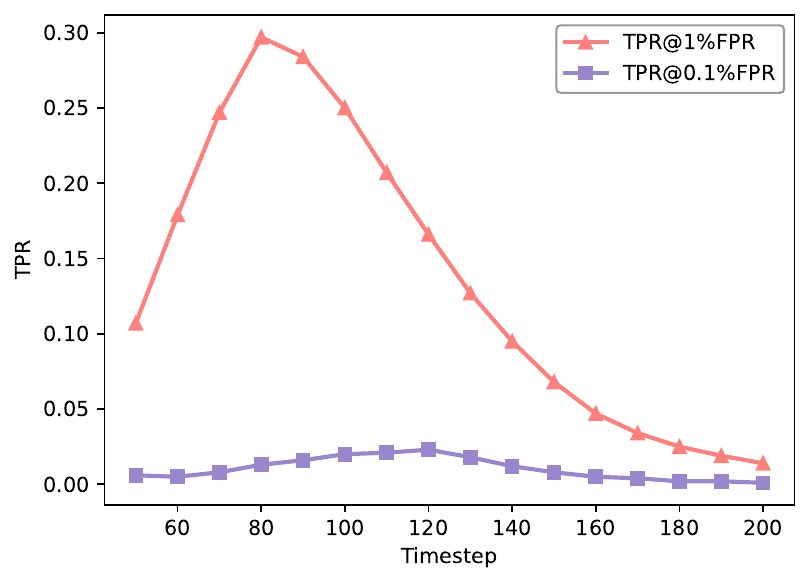}
            \caption{TPR}
            \label{fig:timestep_tpr_fpr}
        \end{subfigure}
        \caption{Attack performance across timesteps}
        \label{fig:timestep}
    \end{minipage}
    \centering
    \begin{minipage}[b]{0.49\textwidth}
        \centering
        \begin{subfigure}[b]{0.49\textwidth}
            \centering
            \includegraphics[width=\textwidth]{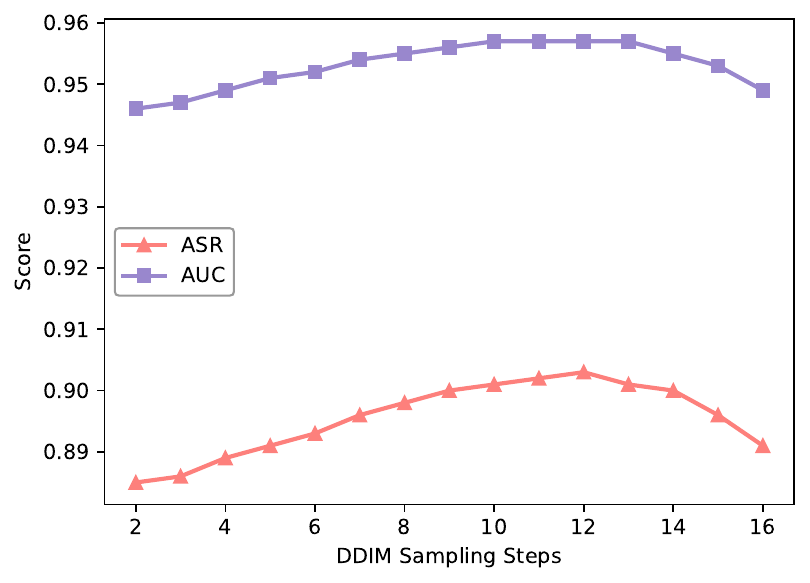}
            \caption{ASR and AUC}
            \label{fig:ddim_asr_auc}
        \end{subfigure}
        \hfill
        \begin{subfigure}[b]{0.49\textwidth}
            \centering
            \includegraphics[width=\textwidth]{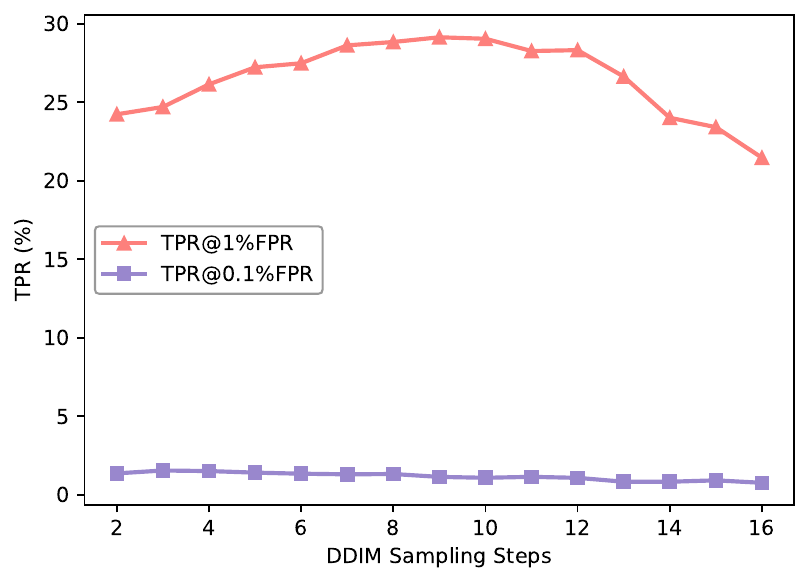}
            \caption{TPR}
            \label{fig:ddim_tpr_fpr}
        \end{subfigure}
        \caption{Attack performance across DDIM sampling intervals}
        \label{fig:ddim_intervals}
    \end{minipage}
    \hfill
\end{figure*}

\subsubsection{Aggregation Degree Metrics Comparative Analysis}

To evaluate the noise aggregation central to our method, we initially employed the L2 mean distance. 
To verify robustness, we explored four alternative metrics: the L1 average distance for outlier resistance, 
the centroid distance for computational simplicity and intuitive concentration measurement, 
the average density for assessing local compactness, 
and the convex hull volume for capturing global geometric dispersion. 
Experimental results (Table~\ref{tab:aggregation_metrics}) indicate that while the L2 mean distance achieves the best performance,
the performance gaps compared to density and centroid-based metrics are negligible, 
confirming that our approach remains robust to the choice of aggregation metric.

\subsubsection{Timestep Parameter Impact Analysis}
We systematically examined the impact of different timesteps on attack
performance. As illustrated in Figure~\ref{fig:timestep}, empirical findings indicate that our method performs well 
within the $T \in [50,140]$ range. We think it is because within this
range, images suffer relatively minimal noise interference and retain sufficient
structural information, enabling diffusion models to effectively predict
original noise based on preserved structural features. In corresponding
neighborhoods, predicted noise distributions for member images are more
concentrated compared to non-member images, 
which enables effective member differentiation through noise aggregation.

\begin{table*}[t] 
\centering
\caption{Performance on Stable Diffusion Models. 
\textbf{Bold} values denote the best results, and \textcolor{blue}{blue} values denote the second-best results.}
\begin{tabular}{@{}lcccccc@{}}
\toprule 
Methods & \multicolumn{3}{c}{SD1.4} & \multicolumn{3}{c}{SD1.5} \\
\cmidrule(lr){2-4} \cmidrule(lr){5-7}
& ASR$\uparrow$ & AUC$\uparrow$ & TPR@1\%FPR$\uparrow$ & ASR$\uparrow$ & AUC$\uparrow$ & TPR@1\%FPR$\uparrow$ \\
\midrule 
GAN-Leaks & 0.533 & 0.468 & 1.73 & 0.541 & 0.472 & 1.64 \\
NaiveLoss & \textcolor{blue}{0.630} & \textcolor{blue}{0.638} & \textbf{23.7} & \textcolor{blue}{0.631} & \textcolor{blue}{0.639} & \textbf{23.7} \\ 
SecMI & 0.602 & 0.605 & \textcolor{blue}{15.3} & 0.602 & 0.606 & \textcolor{blue}{15.3} \\
Ours & \textbf{0.701} & \textbf{0.652} & 8.0 & \textbf{0.696} & \textbf{0.661} & 8.3 \\
\bottomrule
\end{tabular} 
\label{tab:stable_diffusion}
\end{table*}

\subsubsection{DDIM Sampling Intervals Impact Analysis}
We further analyzed the impact of DDIM sampling intervals on overall performance.
As illustrated in Figure~\ref{fig:ddim_intervals},
experiments shows that increasing sampling intervals can enhance attack
performance, but performance stabilizes after reaching a certain range.
We believe the reason is that if the sampling intervals is too short, 
the images will be too similar to identify. 
Both member and non-member images will predict similar noise within adjacent denoising steps, 
making it difficult to distance them. 
But if the sampling intervals is too long, 
the difference between the image and its denoised neighboring images will be too large. 
Even adjacent images denoised by the member image may be considered different images, 
resulting in large differences in noise predictions, which reduced the accuracy of the attack model.
Ultimately, we select sampling interval $m=10$ to balance performance and computational efficiency. 

\subsection{Large-Scale Text-to-Image Model Evaluation}
To validate the effectiveness of our method in practical application scenarios,
we conducted attack testing on mainstream text-to-image
generation models Stable Diffusion v1.4 and v1.5 provided by HuggingFace. 
We randomly sampled 1,000 images from the LAION-aesthetic-5plus dataset as the member
set and sampled 1,000 images from the COCO2017-Val dataset as the non-member
set.

As shown in Table~\ref{tab:stable_diffusion}, our method outperforms existing approaches in terms of ASR and AUC, 
but the attack effectiveness decreases noticeably compared with DDPM. 
We attribute this to the VAE-based latent diffusion process, 
where encoding images into a lower-dimensional space reduces information diversity and 
weakens our method’s ability to distinguish member samples.
Moreover, the large-scale training dataset enhances the model’s generalization capability, 
making it more resistant to MIA.
In addition, our method performs significantly worse than the baseline NaiveLoss~\cite{matsumoto2023membership} on TPR@1\%FPR. 
We attribute this to the fact that the method uses the loss value as the decision criterion, 
making it less affected by model components such as the VAE encoder. 
In particular, when the loss differences are minimal,
the method can confidently infer that the discrepancies arise from training overfitting, 
thereby achieving high-confidence inference at low false positive rates.

\section{Conclusion}
Existing membership inference attack methods for diffusion models typically rely on 
either the loss function or image-level reconstruction comparisons, 
but they struggle to achieve both high accuracy and high efficiency. 
In this work, we propose a membership inference approach based on the aggregation analysis of predicted noises, 
which effectively mitigates the impact of stochastic noise sampling and enables rapid diffusion of samples 
through single-step low-intensity noise injection. 
Our method fully leverages the consistency of predicted noises in local timesteps of diffusion models, 
allowing it to accurately distinguish member samples from non-member samples and substantially reducing the number of model queries.
Empirical findings indicate that our method outperforms existing methods 
on both standard diffusion models, e.g., DDPM and text-to-image models, e.g., Stable Diffusion.
In addition, our analysis of noise intensity and sampling steps provides further insights into the memorization behavior of diffusion models.

\bibliographystyle{IEEEtran} \bibliography{myref}

\end{document}